%% file: main.tex
\documentclass[10pt,twocolumn,letterpaper]{article}

\usepackage[pagenumbers]{wacv} 
\usepackage{amsmath}
\usepackage{booktabs,tabularx, array}
\usepackage{algorithm,algpseudocode, algorithmicx}
\usepackage{siunitx} 
\usepackage{threeparttable}
\usepackage{multirow}
\usepackage{float}
\usepackage[skip=3pt]{caption} 
\usepackage[accsupp]{axessibility} 

\definecolor{wacvblue}{rgb}{0.21,0.49,0.74}
\usepackage[pagebackref,breaklinks,colorlinks,allcolors=wacvblue]{hyperref}

\def\wacvPaperID{*****} 
\def\confName{WACV}
\def\confYear{2027}

\title{Staged Contrastive Pretraining of Heterogeneous Pathology Cohorts: Aligning H\&E with IHC, Transcriptomics, and Proteomics 
}

\author{Olaide N. Oyelade\\
Department of Computer Systems Technology, North Carolina A\&T State University,\\
Greensboro, NC 27411, USA\\
{\tt\small onoyelade@ncat.edu}
\and
Hui Wang\\
School of Electronics, Electrical Engineering and Computer Science, Queen’s University\\
Belfast, United Kingdom\\
{\tt\small h.wang@qub.ac.uk}
}

\begin{document}
\maketitle
\input{sec/0_abstract}
\input{sec/1_intro}

\input{sec/2_relatedworks}
\input{sec/3_methodlogy}
\input{sec/5_result}
\input{sec/6_conclusion}

{
    \small
    \bibliographystyle{ieeenat_fullname}
    \bibliography{main}
}

\end{document}

%% file: sec/0_abstract.tex
\begin{abstract}
Accurate HER2 assessment guides breast cancer therapy, yet most deep learning approaches rely on a single stain or modality and a single cohort. The motivation of the study is driven by the need to understand the implications of both transcriptomics \& proteomics data in informing multimodal contrastive learning of heterogeneous pathology. In addition, we seek to know if staged pairing of multimodal data can improve HER2 scoring since HER2-low matters clinically and H\&E-only scoring is hard. We aim to leverage this multimodal contrastive learning to improve tumor classification, HER2 scoring, and typing of expressed protein. We present a multimodal self-supervised pretraining framework that learns a shared representation from paired H\&E and IHC images, transcriptomics, and proteomics. Because no public dataset provides all four modalities for the same patients, we pretrain on three complementary sources: ACROBAT (H\&E–IHC), TCGA-BRCA (H\&E–RNA), and CPTAC-BRCA (H\&E–protein), in a staged learning approach through warm starts with incremental modality pairing. A presence-masked loss handles modalities that are missing from each sample, so partially paired data can be used without imputation. Our contribution is the combination of cross-source, missing-modality-aware multimodal pretraining with pathology-specific downstream tasks, not a new contrastive objective.
\end{abstract}

%% file: sec/1_intro.tex
\section{Introduction}
\label{sec:intro}
\label{sec:introduction}

Computational pathology increasingly seeks to connect tissue morphology with the molecular characteristics of cancer. Hematoxylin-and-eosin (H\&E) whole-slide images (WSIs) reveal tissue architecture and cellular morphology, while immunohistochemistry (IHC) measures the expression of selected protein markers, transcriptomics describes gene-expression profiles, and proteomics characterises protein abundance. These modalities provide complementary views of tumour biology, but are rarely collected together for every patient. In breast cancer, the assessment of human epidermal growth factor receptor 2 (HER2) is an important example: distinguishing HER2 expression categories, including clinically relevant HER2-low disease, can inform treatment decisions~\cite{modi2022her2low}. Although deep learning can identify correlates of HER2 status from H\&E images~\cite{farahmand2022,lu2022slidegraph}, morphology alone does not directly measure the molecular markers used in pathological scoring.

Learning from complementary assays is complicated by \emph{heterogeneous, partially paired cohorts}. ACROBAT provides H\&E--IHC images from corresponding tissue material~\cite{weitz2024acrobat}; TCGA-BRCA contains pathology images and transcriptomic data~\cite{tcga2012}; and CPTAC supplies breast-cancer proteogenomic resources~\cite{krug2020}. These resources provide different observed associations rather than a single collection of fully matched H\&E, IHC, RNA and proteomic records. Restricting training to complete four-modality samples would therefore exclude useful data. Conversely, assuming that measurements from different patients or cohorts form positive pairs would create spurious supervision. The problem is not merely one of randomly missing inputs: modality availability is correlated with cohort, acquisition protocol and patient population. Moreover, registered H\&E--IHC regions and slide-to-bulk-omics pairs operate at different spatial scales and cannot be treated as equivalent evidence.

Recent H\&E--IHC methods demonstrate the value of cross-stain information for HER2 assessment and transferable representations~\cite{shi2025wsmcl,zhang2025crossstain,oyelade2026}. Histology--omics models further show that molecular supervision can enrich visual representations~\cite{wang2025mirror,robinet2025morpheus,li2026fuji}. However, transferring these ideas to disjoint collections of H\&E--IHC, H\&E--RNA and H\&E--protein pairs requires a training formulation that respects which associations are actually observed. The motivating question of this study is therefore whether these heterogeneous pairings can contribute to a common pathology representation without requiring every modality to be present for every case.

We investigate \emph{staged, presence-masked contrastive pretraining} with H\&E serving as a visual anchor across cohorts. Modality-specific encoders and projections feed a shared fusion architecture. Warm-started stages introduce available H\&E--IHC, H\&E--RNA and H\&E--protein associations, and a modality-presence mask restricts the training objectives to observed inputs and valid positive pairs. The approach is informed by the Contrastive MultiModal (CoMM) framework, which learns multimodal representations by contrasting augmented multimodal views~\cite{dufumier_castillo2025}. Our setting differs in a key respect: the training data do not generally include complete four-way tuples, so the objective must be applied to the available modality combinations. Such training may support a common representation, but does not by itself demonstrate that higher-order interactions among unobserved modality combinations have been learned.

The pretrained model is subsequently examined through four-class HER2 scoring, protein-marker classification and tumour-region detection, following the multi-task design in the accompanying poster. The initial experiments report improving batch-level cross-view retrieval and decreasing pretraining loss, with exploratory downstream metrics. These observations are preliminary and do not yet establish performance gains over appropriate baselines or generalisation to independent cohorts. The study's proposed contributions are: (i) an H\&E-anchored formulation for learning from heterogeneous, partially paired pathology resources; (ii) staged, presence-aware training inspired by CoMM; and (iii) an initial multi-task evaluation of the resulting representations. Whether the staging and additional molecular views improve performance requires controlled ablations and patient-disjoint evaluation.

%% file: sec/2_relatedworks.tex
\section{Related Works}
\label{sec:related_work}

\textbf{HER2 assessment and cross-stain learning:} Deep learning methods have estimated HER2 status from H\&E tumour regions~\cite{farahmand2022} or slide-level graphs~\cite{lu2022slidegraph}. Approaches using complementary stains include weakly supervised H\&E--IHC contrastive learning for HER2 scoring~\cite{shi2025wsmcl}, flexible single- or dual-modality prediction with reconstructed features~\cite{qin2025flexible}, and cross-stain pretraining for transferable WSI representations~\cite{zhang2025crossstain}. Jeong \emph{et al.}~\cite{jeong2026scoreaware} instead investigate HER2-aware virtual IHC through non-contrastive image translation. Of particular relevance, Oyelade \emph{et al.}~\cite{oyelade2026} previously combined H\&E and IHC for HER2 scoring and tumour localisation. The present study extends the research question from paired-stain prediction to pretraining with IHC \emph{and} molecular omics across separate cohorts; its added value must be tested rather than inferred from the number of included modalities. \textbf{Joint histology--omics representations:} MIRROR~\cite{wang2025mirror} balances H\&E--transcriptomic alignment with retention of modality-specific information. MORPHEUS~\cite{robinet2025morpheus} uses masked omics modelling to support variable input subsets and molecular prediction. STORM~\cite{xiang2026storm} investigates spatial transcriptomics--histology alignment, while Fuji~\cite{li2026fuji} uses staged visual pretraining followed by transcriptomic and mutation-based supervision. These are important precedents: neither multimodal molecular grounding nor staging alone is novel. The distinction here is the proposed use of separate H\&E--IHC, H\&E--RNA, and H\&E--proteomics cohort pairings, without presuming cross-cohort molecular measurements refer to the same patient. In particular, spatially localised stain correspondences and bulk-omics associations offer different forms of supervision. \textbf{Contrastive pretraining with incomplete modalities:} CoMM~\cite{dufumier_castillo2025} moves beyond conventional cross-modal pairwise alignment by contrasting augmented joint representations and seeks to preserve shared, unique, and synergistic information. PRIME~\cite{yu2026prime} directly tackles incomplete multimodal cancer cohorts through prototype-based semantic imputation and consistency objectives. The proposed approach adopts a different, more restricted strategy: it warm-starts successive modality combinations and applies contrastive terms only when their required observations are present, without assuming a fully observed four-way sample. This can prevent invalid positive pair construction, but does not automatically resolve cohort confounding, ensure retention of modality-specific biology, or establish four-modality synergy. These limitations motivate comparisons with non-staged training, individual H\&E--omics branches, and existing missing-modality methods, together with cross-cohort and patient-level generalisation tests.

%% file: sec/3_methodlogy.tex
\section{Methodology}
\label{sec:3_methodology}
\subsection{Overview}
\begin{figure*}[t]
  \centering
  \begin{subfigure}{0.65\linewidth}
    \centering
    \includegraphics[width=\linewidth]{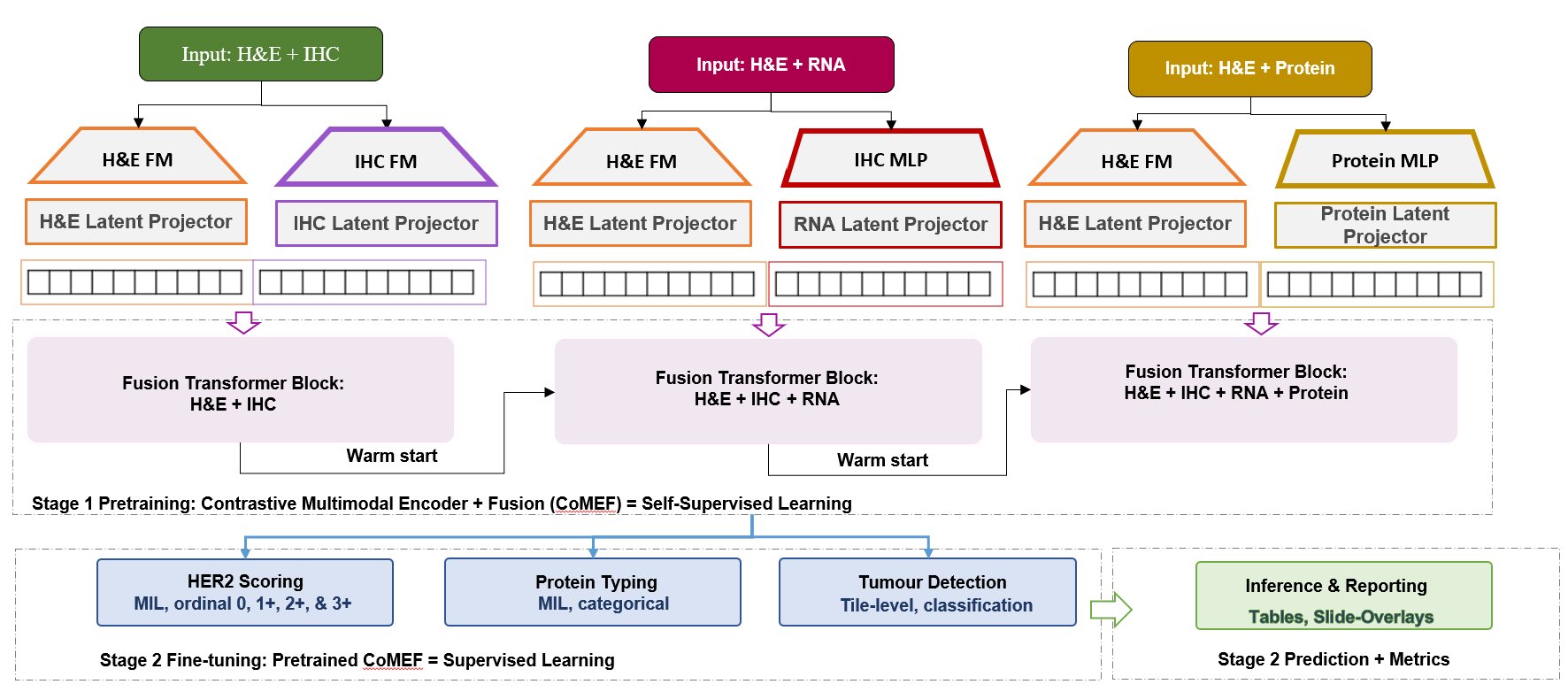}
    \caption{Staged multimodal contrastive learning}
    \label{fig:pipeline}
  \end{subfigure}
  \hfill
  \begin{subfigure}{0.32\linewidth}
    \centering
    \includegraphics[width=\linewidth]{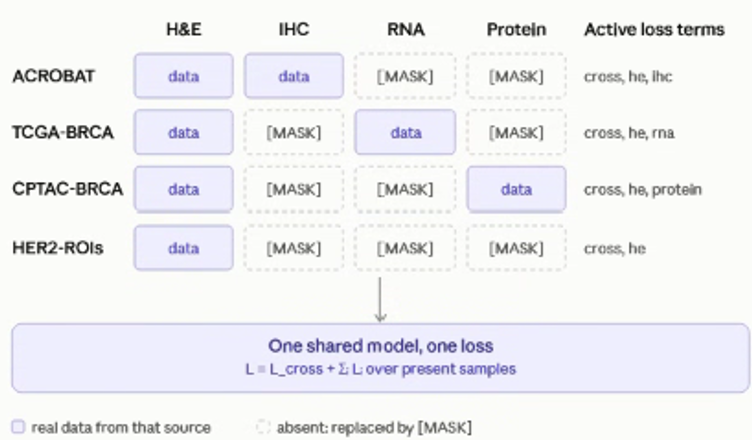}
    \caption{Multimodal dataset sources and integration.}
    \label{fig:data-loss-distribution}
  \end{subfigure}
  \caption{Staged Contrastive Pretraining of Heterogeneous Pathology Cohorts: Aligning
H\&E with IHC, Transcriptomics, and proteomics. (a) shows the end-to-end staged pipeline, while subfigure (b) shows the multimodal data distribution for computing the pre-training loss.}
  \label{fig:full_pipeline_data}
\end{figure*}

Figure \ref{fig:full_pipeline_data} shows the comprehensive architecture of our proposed framework, which incorporates a Presence-Masked Contrastive Learning (PMCL) with unpaired Omics data. This framework uses a three-arm design, and the staged pretraining assumes a contrastive multimodal learning style. The first stage is focused on pre-training on a combination of paired H\&E and IHC samples on a self-supervised learning task. These partially learnt weights are then used to warm-start the second stage, which pairs two other multimodal samples (H\&E and RNA) while preserving the information learnt from IHC. In the third stage, we pivot the degree of multimodality to four by warm-starting the weights of the previous modalities (H\&E, IHC, and RNA) to now include proteomics data. Multimodal samples for this proposed method were drawn from a combined ACROBAT, TCGA, CPTAC, and Yale datasets to develop pair correlations across the four modalities because no single cohort has all four modalities. ACROBAT has H\&E and IHC, TCGA has H\&E and RNA, and CPTAC has H\&E and protein. As illustrated in \ref{fig:pipeline}, the staged pipeline handles cross-view pairs sequentially. Conversely, \ref{fig:data-loss-distribution} maps the underlying multimodal profile cohorts.

In the first stage, which is the full self-supervised phase, the model leverages foundation models (FMs) and our own multilayer perceptron (MLP) for embedding derivation from all multimodal inputs. For instance, for the H\&E / IHC modalities, the study leverages established FMs such as UNI ~\cite{chen2024uni, lu2024avisionlanguage, Lu_2023_CVPR} since these present a strong basis for learnt embeddings given their exposure to $\ge 100k$ H\&E WSIs across 20 tissue types and IHC WSIs. For the omics data embedding learning, we pre-train two separate MLP architectures and integrate them into the framework. As a result, generated embeddings from all modalities were projected into a unified size of shared dimension $d$ using the modality-specific Latent Projector to obtain $\ell$-normalised embeddings $\in\mathbb{R}^{256}$. In Stage 2, we fine-tuned the pre-trained model from Stage 1 for task-specific learning - basically HER2 scoring, protein expression classification, and tumor classification. Meanwhile, in the third stage of the framework, inference on unseen H\&E and IHC slides was applied.   

\subsection{Joint Pretraining (Self-Supervised)}
During the unified self-supervised pretraining phase, the global contrastive objective optimization criterion optimizes alignments across all four modalities simultaneously:
\begin{equation}
    \mathcal{L} = \mathcal{L}_{\text{cross}} + \sum_{i \in \{\mathit{he}, \mathit{ihc}, \mathit{rna}, \mathit{protein}\}} \mathcal{L}_i
\end{equation}

where each sub-modality objective constraint $\mathcal{L}_i$ is computed strictly over samples present within a subset $\mathcal{P}_i$ in a batch. Considering the staged learning applied in this study for multimodal contrastive learning, an $\mathcal{L}_i$ term is dynamically omitted for any given batch if the active subset does not contain the $\mathcal{L}_i$ modality and the count for a modality falls below two:
\begin{equation}
    \mathcal{L}_i = \begin{cases}
        \text{Contrastive Loss} & \text{if } |\mathcal{P}_i \cap \mathcal{B}| \ge 2, \\
        0 & \text{otherwise.}
    \end{cases}
\end{equation}
where $\mathcal{B}$ denotes the current batch. However, since our multimodal framework handles all modalities across the staged learning, this provides an opportunity to achieve cross-modality, synergistic, unique, and redundant pattern learning across modalities ~\cite{dufumier_castillo2025}. Meanwhile, to achieve joint learning of H\&E/IHC across heterogeneous multi-omics repositories without requiring fully matched profiles, a case-level presence subset $\mathcal{P}_m$ for each modality $m \in \{\mathit{he}, \mathit{ihc}, \mathit{rna}, \mathit{protein}\}$, then we define modality present set combination as $\mathcal{P}_{\mathit{he}} = \{t: t \text{ is any tile across all four data sources}\}$, leading to $\mathcal{P}_{\mathit{ihc}} = \{t : t \in \text{ACROBAT} \text{ and has a successfully registered IHC pair}\}$, $\mathcal{P}_{\mathit{rna}} = \{t : t \in \text{TCGA-BRCA}\}$ and $\mathcal{P}_{\mathit{protein}} = \{t : t \in \text{CPTAC-BRCA}\}$. Using the Virtual Alignment of pathoLogy Image Series (VALIS) ~\cite{gatenbee2023valis}, we register H\&E and IHC WSIs using rigid/non-rigid transformations. The multimodal image and omics registration is followed by tiling of the H\&E/IHC from ACROBAT supplies paired for the cross-stain alignment pretext task learning. Registration/tiling was also done for the  H\&E/RNA-seq plus (noisy) TCGA supplies (used for HER2 status), and H\&E/real proteomics from CPTAC supplies. For each modality, we generated two augmentations $\text{aug}_1,\text{aug}_2$ ($X'$ and $X''$,) to provide the model with a positive pair that actually are rather than treating the modalities (H\&E and IHC or H\&E and RNA) as the two views replacing augmentations. The embedding generated for every single modality ($Z_m$) will be paired with one of the view augmentations ($Z'$). 

The fusion model ($f$) and its loss functions are preceded by each modality encoder $F_m$ and a per-modality latent converter (shared embedding projection $d$). In addition to this input ($x_m$) token sequence ($F_m$), most critically, missing modalities in the input are replaced with a learned mask token $\mathbf{m}$ whenever that modality is absent or intentionally masked for a given sample.
$$h_m = \begin{cases} \text{Converter}_m(F_m(x_m)) & \text{if present} \\ \mathbf{m} & \text{if masked or absent} \end{cases}$$

The fusion transformer ($f_m$) concatenates a [CLS] token with every modality's token sequence and applies self-attention across the whole sequence so that the encoded [CLS] output is the learned fused embedding.
$$Z = f\big([\text{CLS}] \,\|\, h_{he} \,\|\, h_{ihc} \,\|\, h_{rna} \,\|\, h_{protein}\big)_{[\text{CLS}]}$$

Going by the $Z$, the $f$ is able to take the augmented views of the same multimodal sample ($X'$ and $X''$) to produce $Z'$ and $Z''$. The cross-view multimodality loss is computed using:

$$\mathcal{L}_{\text{cross}} = -\log\frac{\exp(\text{sim}(Z',Z'')/\tau)}{\sum_{k}\exp(\text{sim}(Z',Z_k'')/\tau)}$$

For each modality $i$, a single-modality view, while keeping every other modality masked, produces $Z_i$. This allows for preserving the unique signal from each modality during the fusion process learning. This penalizes leaning heavily on dominant/easier modalities so that other modalities are not silently discarded. Therefore, the $m_i$ are contrasted against both full views, yielding:

\begingroup 
\setlength{\abovedisplayskip}{2pt}
\setlength{\belowdisplayskip}{2pt}
\begin{align}
\mathcal{L}_i = \frac{1}{2}\left[I(Z_i, Z') + I(Z_i, Z'')\right]
\end{align}
\endgroup
This leads to the total loss:
\begin{align}
\mathcal{L} = \mathcal{L}_{\text{cross}} + \sum_{i \in \{he,ihc,rna,protein\}} \mathcal{L}_i
\end{align}

Given $n$ samples in a batch containing a specific modality with latent embeddings $u_k$ and $v_l$, the scaled cosine similarity $s_{kl}$ is described as:
\begingroup 
\setlength{\abovedisplayskip}{4pt}
\setlength{\belowdisplayskip}{4pt}

\begin{equation}
    s_{kl} = \frac{\langle u_k, v_l \rangle}{\|u_k\| \|v_l\| \tau}
\end{equation}
\endgroup
For the cross-view global objective $\mathcal{L}_{\text{cross}}$, we assign $u = Z'$ and $v = Z''$. The directional row and column components tracking individual sample alignment are described as:
\begin{align}
    r_k &= -\log \left[ \frac{\exp(s_{kk})}{\sum_{l=1}^{n} \exp(s_{kl})} \right] \\
    c_l &= -\log \left[ \frac{\exp(s_{ll})}{\sum_{k=1}^{n} \exp(s_{kl})} \right]
\end{align}
The two directed symmetric losses are computed as:
\begin{align}
    \ell(Z', Z'') &= \frac{1}{n} \sum_{k=1}^{n} r_k = -\frac{1}{n} \sum_{k=1}^{n} \log \left[ \frac{\exp(s_{kk})}{\sum_{l=1}^{n} \exp(s_{kl})} \right] \\
    \ell(Z'', Z') &= \frac{1}{n} \sum_{l=1}^{n} c_l = -\frac{1}{n} \sum_{l=1}^{n} \log \left[ \frac{\exp(s_{ll})}{\sum_{k=1}^{n} \exp(s_{kl})} \right]
\end{align}
The per-modality (others masked out) is defined as:
\begin{align}
    \mathcal{L}_i &= \tfrac{1}{2} \left[ \ell(u,v) + \ell(v,u) \right]
\end{align}

The Algorithm~\ref{alg:ssl} details the self-supervised learning phase described in this subsection.

\begin{algorithm}[t]
\caption{Multimodal Self-Supervised Pretraining}
\label{alg:ssl}
\begin{algorithmic}[1]
\Require Tile manifest $\mathcal{D}$; modalities $\mathcal{M} = \{\mathit{he}, \mathit{ihc}, \mathit{rna}, \mathit{prot}\}$; frozen foundation image encoders $f_{\mathit{he}}, f_{\mathit{ihc}}$; trainable omics encoders $f_{\mathit{rna}}, f_{\mathit{prot}}$; linear projections $\pi_m$; learned mask tokens $\mu_m$; fusion transformer $g$ with $[\text{CLS}]$ token; contrastive temperature $\tau$, minibatch size $B$, learning rate $\eta$.
\For{\textbf{epoch} $= 1, \dots, E$}
    \For{\textbf{each minibatch} $\{x_b\}_{b=1}^{B} \subset \mathcal{D}$}
        \State $p_{b,m} \in \{0, 1\}$ \Comment{\small $m$ modality present $b$}
        \State $\mathcal{M}_{\text{batch}} \leftarrow \big\{m \in \mathcal{M} : \sum_b p_{b,m} > 0\big\}$ 
        
        \State \textbf{Construct Token Embeddings:}
        \For{$b = 1, \dots, B$ \textbf{and} $m \in \mathcal{M}$}
            \If{$p_{b,m} == 1$}
                \State $t_{b,m} \leftarrow \pi_m\big(f_m(x_{b,m})\big)$
            \Else
                \State $t_{b,m} \leftarrow \mu_m$
            \EndIf
        \EndFor
        
        \State \textbf{Multi-View Fusion Passes:}
        \State $Z' \leftarrow g\big(\text{aug}_1(t)\big)_{[\text{CLS}]}, \quad Z'' \leftarrow g\big(\text{aug}_2(t)\big)_{[\text{CLS}]}$
        
        \State \textbf{Single-Modality Alignment Passes:}
        \For{$m \in \mathcal{M}_{\text{batch}}$}
            \State $\hat{t}^{(m)}_{b} \leftarrow \big\{t_{b,m} \text{ if } \text{target modality } m, \text{ else } \mu\big\}$
            \State $Z_m \leftarrow g\big(\hat{t}^{(m)}\big)_{[\text{CLS}]}$
        \EndFor
        
        \State \textbf{Objective Criteria Optimization:}
        \State $\mathcal{L}_{\text{cross}} \leftarrow \mathcal{L}_{\mathrm{NCE}}(Z', Z''; \tau)$
        \For{$m \in \mathcal{M}_{\text{batch}}$}
            \State $S_m \leftarrow \{b : p_{b,m} == 1\}$
            \If{$|S_m| \ge 2$}
                \State $\mathcal{L}_m \leftarrow \mathcal{L}_{\mathrm{NCE}}\big(Z_m[S_m], Z'[S_m]; \tau\big)$
            \Else
                \State $\mathcal{L}_m \leftarrow 0$ \Comment{\small Insufficient valid subset samples}
            \EndIf
        \EndFor
        
        \State $\mathcal{L} \leftarrow \mathcal{L}_{\text{cross}} + \sum_{m \in \mathcal{M}_{\text{batch}}} \mathcal{L}_m$
        
        \State \textbf{Parameter Step Update:}
        \If{$\mathcal{L}$ \textbf{and} $\nabla_\theta\mathcal{L}$ are finite}
            \State Clip gradients $\nabla_\theta\mathcal{L}$
            \State $\theta \leftarrow \mathrm{AdamW}(\theta, \nabla_\theta\mathcal{L}, \eta)$
        \Else
            \State \textbf{skip step}
        \EndIf
    \EndFor
    \State Save checkpoint (weights $\theta$)
\EndFor
\State \Return $\theta$
\end{algorithmic}
\end{algorithm}

\subsection{Multi-Task Supervised Fine-Tuning}
With the pre-trained model completed, the downstream fine-tuning phase depends on labeled data for task-specific supervision. This is aimed at optimizing the pretrained weights for HER2 scoring, tumor classification, and typing of protein expression. Considering the existing data distribution, for the HER2 protein expression classification task, labels were generated from the TCGA-BRCA and HER2 Tumor ROIs Subset. The ACROBAT Multistain dataset was sourced for the stain type labels for HER2 scoring, and the HER2 Tumor ROIs, Yale Cohorts provided samples for deriving labels for tumor classification. The combined multi-task fine-tuning loss $\mathcal{L}_{\text{ft}}$ balances these three multitasks using scalar hyperparameters $\lambda_1, \lambda_2, \lambda_3$:
\begin{equation}
    \mathcal{L}_{\text{ft}} = \lambda_{\text{her2}}\,\mathcal{L}_{\text{her2}} + \lambda_{\text{protein}}\,\mathcal{L}_{\text{protein}} + \lambda_{\text{tumor}} \cdot \frac{1}{N}\sum_{i=1}^{N}\mathcal{L}_{\text{tumor},i}
\end{equation}

where $\mathrm{L}$ denotes the standard cross-entropy loss criterion, and $\mathrm{N}$ is the number of tiles in the WSI. Since the HER2 scoring and protein typing tasks are WSI-level classification tasks, a multi-task head Attention-MIL pooling learns a per-tile weight $\alpha_i$ as seen here:

\begin{equation}
\alpha_i = \frac{\exp\big(w^\top \tanh(V h_i)\big)}{\sum_{j=1}^{N}\exp\big(w^\top \tanh(V h_j)\big)}, \qquad z_{\text{slide}} = \sum_{i=1}^{N} \alpha_i h_i
\end{equation}

Since HER2 scoring is an ordinal scale $(0 < 1+ < 2+ < 3+)$ classification task, we used the quadratic-weighted Cohen's kappa,  a metric that allows a 0-vs-3+ to be penalized far more than 0-vs-1+. Similarly to HER2 scoring, the protein-typing logits, a linear projections of $z_{\text{slide}}$ is trained with standard cross-entropy $\mathcal{L}(y,\hat{y}) = -\sum_c y_c \log \hat{y}_c$. On the other hand, the tumor classifier is a linear projection applied directly to every $h_i$ (tiles in $z_{\text{slide}}$), producing $N$ independent per-tile predictions. The combined fine-tuning objective sums all three, with a sample skipped from a given term entirely when it lacks that label. This implies that a HER2-only slide will skip protein loss, while a H\&E-only slide will skip nothing since the fusion transformer's mask token already handles missing IHC/omics upstream.

\subsection{Inference}
During the inference phase, we omit the dual view augmentation described in the self-supervised stage and instead input a slide's tiles. Whereas the pre-training and fine-tuning phases allow for unfreezing the encoders, during inference, we keep these encoders and the fusion transformer frozen simply to generate an embedding ($Z$) per tile. The MIL pooling produces one HER2 score and one protein type for every given WSI input, while the tumor classifier produces per-tile classes/categorization. These tile-level classifications are then combined back onto the slide thumbnail as a spatial overlay.

%% file: sec/5_result.tex
\section{Experimentation, Results, and Discussion}
All training was carried out in a computational environment with $Nvidia RTX 6000$ Ada Generation, $48 GB$ GDDR6, $64GB$, memory, Intel Xeon $w7-3565X$, and a 2 TB performance SSD, SED-ready, units based on the $A100$ Tensor Core GPU with a memory size of $48 GB$.   
\subsection{Hyperparameter configuration}
A comprehensive architecture configuration, modality-specific encoder traits, and structural hyperparameter thresholds used to train the proposed framework are described here. For the image (H\&E and IHC) encoder types, such as the foundation models (FMs), a feature-token architecture was used for the RNA and protein (omics) data. Input dimension for the RNA and protein were held at 2000 and 1500 respectively by finding only features with marginal variance for elimination while retaining others, while maintaining a uniform embedding dimension of 256 for both, and the hidden dimension is 256. On the other hand, the embedding dimension of 1024 was used for both H\&E and IHC. Backbone encoders (FMs) for the WSIs were unfrozen to improve domain-specific learning based on the datasets. For the fusion transformer, we used a token dimension of 256 and maintained a single-layer, \(8\)-headed Fusion Transformer. We experimented with pre-training using a batch size of 64 across the pre-training and fine-tuning epochs. We utilized an AdamW optimizer while holding a learning rate of $3 \times 10^{-4}$, weight decay of $10^{-2}$, and temperature ($\tau$) of $0.1$. The FM-initialised image encoder was trained further during multimodal pretraining, and in the end-to-end fine-tuning it was updated again.

\subsection{Multimodal datasets}
The multimodal datasets used for the experimentation were sourced from publicly available repositories. These include ACROBAT~\cite{weitz2024acrobat}, TCGA-BRCA~\cite{tcga2012,cerami2012,gao2013}, CPTAC-BRCA~\cite{krug2020,tcia2013}, HER2 tumour ROIs~\cite{farahmand2022,farahmand2022tcia}, Yale HER2 cohort~\cite{farahmand2022,farahmand2022tcia} and Yale trastuzumab~\cite{farahmand2022,farahmand2022tcia}. The ACROBAT dataset provided us with two modalities, namely the H\&E and IHC WSIs (with protein + ER/PR/HER2/Ki-67). Although there were 4,212 WSIs from 1,153 breast cancer patients, with up to five stainings per patient, we used 200 WSIs (both H\&E and IHC split across 29 ER, 29 KI67, 28 PGR and 14 HER2) from 100 cases. The registered H\&E and IHC pairs were used for Stage-1 pretext task pretraining. The HER2-TUMOR-ROIS (TCIA), which combines the TCGA-BRCA subset with a Yale cohort, provided us with the H\&E WSIs with their associated tumor ROI annotations and HER2 status modalities, supplying 182 samples (90 HER2- and 92 HER2+). The  Clinical Proteomic Tumor Analysis Consortium Breast Invasive Carcinoma Collection (CPTAC-BRCA) repository provided us with two modalities covering both H\&E and proteomics. The CPTAC-BRCA consists of 642 whole-slide images of 134 patients, and also proteomics data, supplying us with the proteomics paired with H\&E WSI. The datasets and their corresponding sample configurations are summarized in Table ~\ref{tab:datasets}.

\begin{table}[h]  
\centering
\scriptsize      
\begin{threeparttable}
\begin{tabularx}{\linewidth}{@{} l >{\hsize=1.2\hsize\raggedright\arraybackslash}X >{\hsize=0.8\hsize\raggedright\arraybackslash}X r @{}}
\toprule
Dataset & Modalities & Stage & Paired Slides \\
\midrule
ACROBAT~\cite{weitz2024acrobat}\tnote{a}
& H\&E + IHC (HER2, ER, PR, Ki67)
& Stage 1,2, arm 1; 
& {200} \\
\addlinespace
TCGA-BRCA~\cite{tcga2012,cerami2012,gao2013}\tnote{b}
& H\&E + RNA-seq
& Stage 1, 2, arm 2;
& 394 \\
\addlinespace
CPTAC-BRCA~\cite{krug2020,tcia2013}\tnote{c}
& H\&E + proteomics
& Stage 1, arm 3
& 653 \\
\addlinespace
TCIA HER2 tumour~\cite{farahmand2022}\tnote{d}
& H\&E + tumour
& Stage 2
& 273 \\
\addlinespace
Yale HER2~\cite{farahmand2022}\tnote{e}
& H\&E + tumour
& Stage 2
& 188 \\
\addlinespace
Yale trastuz.~\cite{farahmand2022}\tnote{f}
& H\&E + tumour
& Stage 2
& 85 \\
\bottomrule
\end{tabularx}
\caption{Datasets, their sample metrics, and structured pipeline configurations. }
\label{tab:datasets}
\begin{tablenotes}
    \tiny  
    \item[a] \textbf{ACROBAT:} Paired H\&E--IHC alignment for self-supervision; protein-type head for stain identity. Used in pretraining and fine-tuning phases.
    \item[b] \textbf{TCGA-BRCA:} H\&E--RNA alignment; HER2 IHC score (0, 1+, 2+, 3+) derived from cBioPortal. Used in pretraining and fine-tuning phases.
    \item[c] \textbf{CPTAC-BRCA:} H\&E--protein alignment (one slide per patient). Restricted to the pretrain phase only.
    \item[d] \textbf{HER2 tumour ROIs:} Tumour/non-tumour tile labels; binary HER2 status. Used in the fine-tuning phase.
    \item[e] \textbf{Yale HER2 cohort:} Tumour tile labels; binary HER2 status. Used in the fine-tuning phase.
    \item[f] \textbf{Yale trastuzumab-response:} Tumour tile labels; HER2 breast cancer IHC score (0--3). Used in the fine-tuning phase.
\end{tablenotes}
\end{threeparttable}
\vspace{-0.5cm}
\end{table}

Since splitting by slide or tile would leak a patient into both train and test, we approached this by using a cohort for pretraining and a different cohort for inference.
\subsection{ Result and Performance Evaluation}
The results of the experimentation are detailed in this section. Table ~\ref{tab:pretrain_metrics} outlines the results obtained for the unsupervised pretraining phase, observed under 10 epochs. As evaluated in the table, the performance observed from our joint multimodal self-supervised pretraining shows that there was strong model convergence and feature alignment over the pre-training optimization window. The implication of these findings is that, whereas most studies often focus on H\&E and IHC alignment, our study shows a promising potential for strong alignment beyond just those two modalities, but also across the omics data. Furthermore, the result indicates a progressive alignment acceleration since the batch-averaged cross-view retrieval accuracy increases significantly, rising steadily from 0.252 to 0.588. This interesting increase in the cross-view retrieval accuracy confirms that the fusion transformer is successfully structuring a highly stable, shared latent space. Also, there is a clear and continuous minimization of the joint loss function, which is an expected and desirable multiobjective minimization, with the total loss dropping from 7.764 to 6.043. The observed decline was jointly contributed by the cross-View and modality encoders, whose loss scores decreased by 38.9\% and 39.1\%, respectively. Interestingly, we observed that while the pathology images minimized learning loss towards the contrastive constraints, the omics modality (e.g the RNA modality loss) remained steady at approximately 2.109. The implication of this is that the joint self-supervised learning prioritized macro-structural tissue geometry learning before fully mapping to the complex molecular profiles.

\begin{table}[htbp]
\centering
\footnotesize
\begin{tabularx}{\linewidth}{
  @{} c 
  >{\centering\arraybackslash}X 
  >{\centering\arraybackslash}X 
  >{\centering\arraybackslash}X 
  >{\centering\arraybackslash}X 
  >{\centering\arraybackslash}X 
  >{\centering\arraybackslash}X @{}
}
\toprule
& & \multicolumn{5}{c}{Loss Components $\downarrow$} \\
\cmidrule(l){3-7}
Epoch & \parbox[c]{1.2cm}{\centering Retr.\\Acc. $\uparrow$} & \parbox[c]{1.0cm}{\centering Cross-\\View} & \parbox[c]{1.1cm}{\centering Mod.\\H\&E} & \parbox[c]{1.1cm}{\centering Mod.\\IHC} & \parbox[c]{1.1cm}{\centering Mod.\\RNA} & \parbox[c]{1.1cm}{\centering Total\\Loss} \\
\midrule
0 & 0.252 & 1.854 & 1.927 & 1.886 & 2.097 & 7.764 \\
1 & 0.317 & 1.795 & 1.776 & 1.877 & 2.092 & 7.541 \\
2 & 0.348 & 1.739 & 1.689 & 1.842 & 2.094 & 7.363 \\
3 & 0.261 & 1.842 & 1.777 & 1.896 & 2.095 & 7.610 \\
4 & 0.315 & 1.741 & 1.707 & 1.858 & 2.089 & 7.396 \\
5 & 0.396 & 1.581 & 1.484 & 1.850 & 2.098 & 7.014 \\
6 & 0.442 & 1.531 & 1.365 & 1.856 & 2.100 & 6.853 \\
7 & 0.489 & 1.436 & 1.288 & 1.830 & 2.105 & 6.660 \\
8 & 0.516 & 1.314 & 1.307 & 1.706 & 2.109 & 6.435 \\
9 & \textbf{0.588} & \textbf{1.133} & \textbf{1.173} & \textbf{1.628} & 2.109 & \textbf{6.043} \\
\bottomrule
\end{tabularx}
\caption{Multi-modal pretraining evaluation metrics across initial optimization epochs.}
\vspace{-0.4cm}
\label{tab:pretrain_metrics}
\end{table}

\begin{figure}[htbp]  
  \centering
  \begin{subfigure}{0.48\linewidth}
    \centering
    \includegraphics[width=\linewidth]{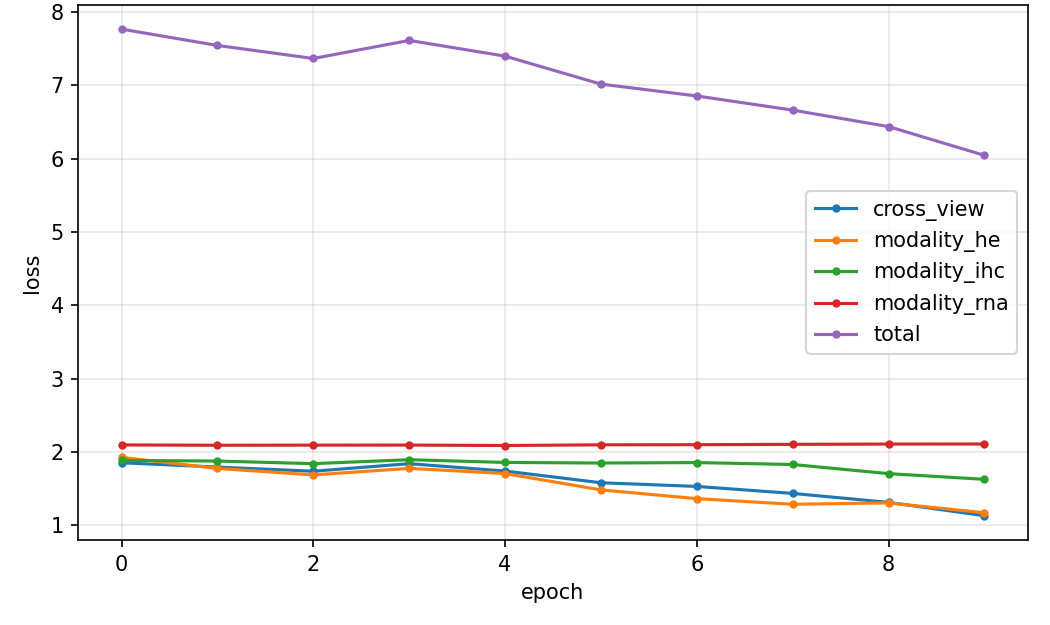}
    \caption{Loss trajectories.}
    \label{fig:pretrain-loss}
  \end{subfigure}
  \hfill 
  \begin{subfigure}{0.48\linewidth}
    \centering
    \includegraphics[width=\linewidth]{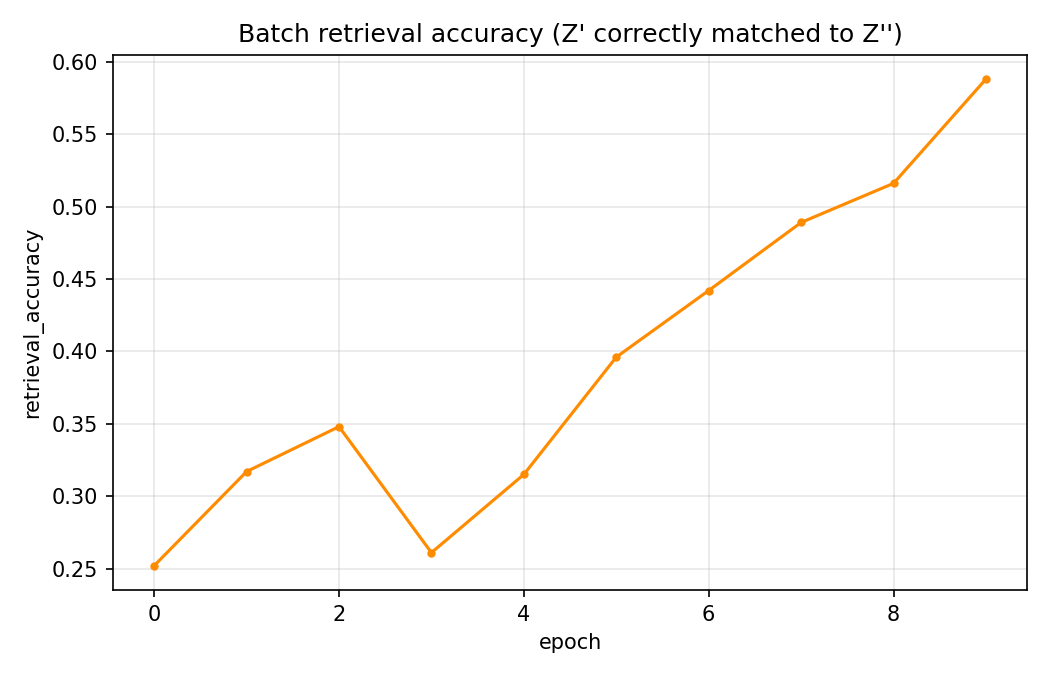}
    \caption{Retrieval accuracy.}
    \label{fig:pretrain-accuracy}
  \end{subfigure}
  \caption{Multimodal pretraining evaluation curves: (a) monitors the lower-bound convergence, cross-view and individual sub-modality encoders, (b) tracks latent feature retrieval alignment.}
  \vspace{-0.6cm}
  \label{fig:pretrain-metrics-curves}
\end{figure}

We plotted the training curves to visualize the alignment of modalities during the self-supervised stage. Figure~\ref{fig:pretrain-metrics-curves} shows a measure of latent feature convergence, with the total loss curve exhibiting a steady downward trajectory from 7.76 to 6.04, confirming stable training optimization. Similarly, the curves show sharp retrieval accuracy gains as seen in Figure~\ref{fig:pretrain-metrics-curves}(b), after an initial exploration dip at epoch 3, the cross-view batch retrieval accuracy shows a sharp increase by epoch 9.

\paragraph{Multi-task Latent Fine-tuning performance} The performance of the second stage of our proposed framework was also evaluated. This multi-task fine-tuning stage is meant to adapt the generalistic self-supervised learning to labels. The results obtained for the first four epochs are listed in Table.~\ref{tab:tumor_metrics}. The result indicates that the tumor-ROI tile classification trajectory shows increased performance gains on 53,800 tile samples, with a classification loss dropping steadily by 15.8\% and accuracy improving up to 79.8\% just within epoch 4. Furthermore, an evaluation of the class metrics shows that there is an upward trend, with macro F1, precision, and recall returning values of 0.749, 0.756, and 0.743, respectively. This is a testament that the fine-tuning learning maps asymmetric tissue boundaries without overfitting the majority class. Interestingly, the quadratic weighted Kappa (\(\kappa \)) metric yields the largest overall gain increase by 26.4\%, a confirmation of the model's prediction errors are tightening. Furthermore, in Table ~\ref{tab:multitask_evaluation}, the results outline showed that our proposed framework achieves strong optimization gains across all three downstream evaluation tasks within a single fine-tuning epoch. For instance, we observed substantial molecular classification, stable clinical target optimization, and better spatial tissue understanding. The multi-omics target of protein profiling shows the largest overall performance jump, with its prediction accuracy increasing by 25.0\%.On the other hand, the HER2 predictor reaches a solid 61.2\% accuracy and a 0.469 Macro F1 just at epoch 1. Meanwhile, the structural task, tumor classification, shows a loss reduction down to 0.467 and classification accuracy up to 76.1\%, while quadratic weighted kappa (\(\kappa \)) improves by 22.1\%.

\begin{table}[htbp]
\centering
\footnotesize 
\begin{tabularx}{\linewidth}{@{} c >{\centering\arraybackslash}X >{\centering\arraybackslash}X >{\centering\arraybackslash}X >{\centering\arraybackslash}X >{\centering\arraybackslash}X >{\centering\arraybackslash}X @{}}
\toprule
Epoch & Loss $\downarrow$ & Acc. $\uparrow$ & \parbox[c]{0.9cm}{\centering Macro\\F1} & \parbox[c]{1.2cm}{\centering Precis.} & \parbox[c]{1.0cm}{\centering Recall} & \parbox[c]{1.1cm}{\centering Kappa\\($\kappa$)} \\
\midrule
0.0 & 0.492 & 0.764 & 0.695 & 0.717 & 0.684 & 0.394 \\
1.0 & 0.441 & 0.786 & 0.729 & 0.743 & 0.719 & 0.459 \\
2.0 & 0.430 & 0.791 & 0.736 & 0.749 & 0.727 & 0.473 \\
3.0 & 0.424 & 0.792 & 0.742 & 0.751 & 0.735 & 0.484 \\
4.0 & \textbf{0.414} & \textbf{0.798} & \textbf{0.749} & \textbf{0.756} & \textbf{0.743} & \textbf{0.498} \\
\bottomrule
\end{tabularx}
\caption{Tumor-ROI Morphological Tile Classification Trajectory. }
\vspace{-0.5cm}
\label{tab:tumor_metrics}
\end{table}

\begin{table}[htbp]
\centering
\scriptsize 
\begin{tabularx}{\linewidth}{@{} l c c >{\centering\arraybackslash}X c c c @{}}
\toprule
Target Domain & Loss $\downarrow$ & Acc. $\uparrow$ & \parbox[c]{0.8cm}{\centering Macro\\F1} & \parbox[c]{1.0cm}{\centering Precis.} & \parbox[c]{1.0cm}{\centering Recall} & \parbox[c]{1.0cm}{\centering Kappa\\($\kappa$)} \\
\midrule
\multicolumn{7}{@{}l}{\textit{Initial Validation Baseline (Epoch 0.0)}} \\
\midrule
Tumor Pathology & 0.515 & 0.743 & 0.668 & 0.684 & 0.659 & 0.339 \\
Protein Profile & 1.484 & 0.280 & 0.192 & 0.183 & 0.242 & 0.091 \\
HER2 Predictor  & 0.829 & 0.541 & 0.314 & 0.318 & 0.314 & 0.070 \\
\midrule
\multicolumn{7}{@{}l}{\textit{Multi-Task Optimization Result (Epoch 1.0)}} \\
\midrule
Tumor Pathology & \textbf{0.467} & \textbf{0.761} & \textbf{0.707} & \textbf{0.712} & \textbf{0.702} & \textbf{0.414} \\
Protein Profile & \textbf{1.411} & \textbf{0.350} & \textbf{0.281} & \textbf{0.266} & \textbf{0.305} & 0.055 \\
HER2 Predictor  & \textbf{0.706} & \textbf{0.612} & \textbf{0.469} & \textbf{0.506} & \textbf{0.503} & 0.007 \\
\bottomrule
\end{tabularx}
\caption{Comparative Multi-Task Evaluation of Downstream Modality Alignment.}
\vspace{-0.3cm}
\label{tab:multitask_evaluation}
\end{table}

\begin{figure}[t] 
  \centering
  \begin{subfigure}{0.48\linewidth}
    \centering
    \includegraphics[width=\linewidth]{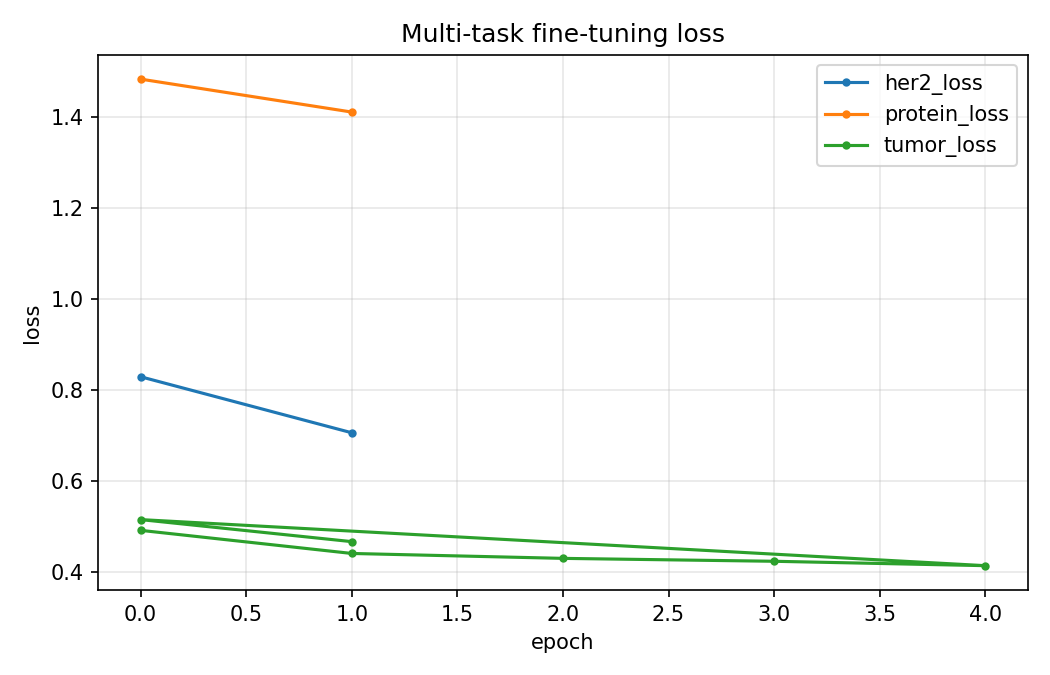}
    \caption{Multi-task loss curves.}
    \label{fig:multitask_loss_curves}
  \end{subfigure}
  \hfill 
  \begin{subfigure}{0.48\linewidth}
    \centering
    \includegraphics[width=\linewidth]{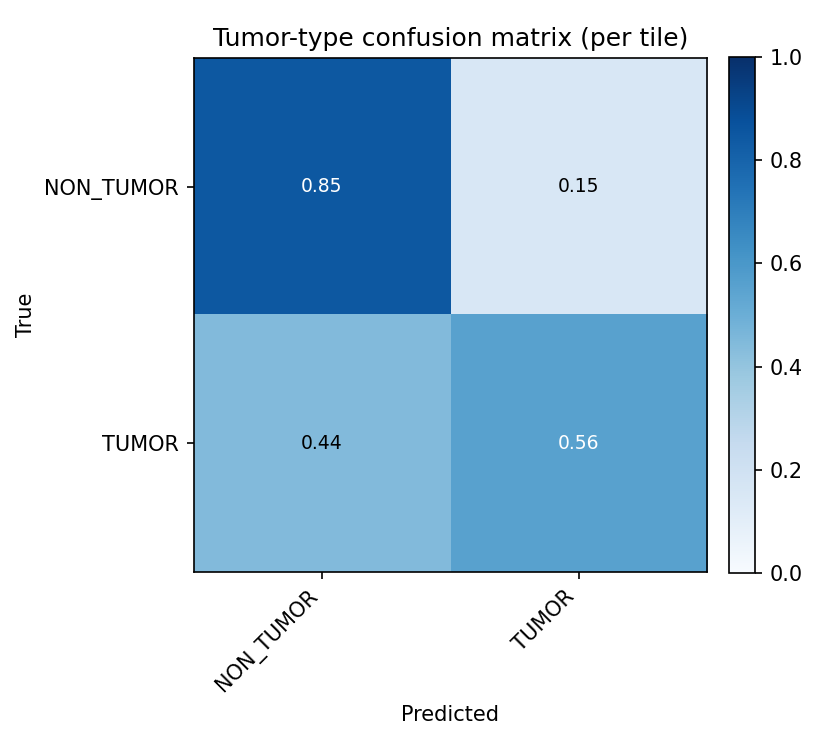}
    \caption{Tumor confusion matrix.}
    \label{fig:tumor-confusion-matrix}
  \end{subfigure}
  \caption{Downstream fine-tuning metrics (a) tracks structural gradient convergence, (b) evaluates tile-level tumor vs. non-tumor}
  \label{fig:fine_tuning_dashboard}
  \vspace{-0.7cm}
\end{figure}

As plotted in Figure ~\ref{fig:multitask_loss_curves}, the multi-task fine-tuning shows a highly effective optimization across heterogeneous diagnostic objectives of efficient cross-task joint learning, biomarker target convergence, and morphological optimization. This shows that the framework demonstrates its capability to handle diverse operational tasks simultaneously. For instance, the protein loss drops steadily from 1.48 to 1.41 in its first epoch, while the core clinical scoring metric, HER2 loss, falls quickly from 0.83 to 0.70 within just one epoch. This stabilization seen among the two losses confirms that co-embedding multi-view spatial slides with molecular features during Stage 1 pretraining yields a latent space that generalizes well to cancer classification and protein profiling. Moreover, the patch-level tumor loss decreased steadily, reaching a lower bound of 0.41. The implication is that, during fine-tuning, the shared model handles both whole-slide and tile-level spatial masking smoothly. In addition, the plot in Figure ~\ref{fig:tumor-confusion-matrix} shows the tile-level structural tissue classification task. During the initial multi-task optimization phase, we observed that the model achieves a strong 85\% true negative accuracy when classifying healthy, non-malignant tissue structures. This high specificity proves that the spatial feature extraction capabilities learned during Stage 1 self-supervised pretraining successfully distinguish benign morphology. On the other hand, we observed that although the true positive detection sits at 56\%, the model still minimizes total failure modes by maintaining an active discriminative slope across the diagonal, while the cross-class contamination rate was 44\%. 

\textbf{Model Inference - performance evaluation} We ran predictions on the three tasks to examine the inference capability of the fine-tuned model. In Table ~\ref{tab:inference_sample_manifest}, we outline the observed performance patterns seen in the raw logs. While the prediction confidence values remain constrained between 0.260 and 0.460 across all 3 sub-task evaluations, we also observed that it over-predicts 3+ for the majority of the TCGA rows while simultaneously defaulting to HER2 for the protein-type predictions. Despite early-stage class bias, the model correctly hits exact cross-modality matches on specific hard test samples (e.g., Case 23, Case 35, and Case 48), where it accurately matches the hidden HER2 and PGR protein types. This performance is indicative of the early training stages whose checkpoint was applied for inference. 

\begin{table}[htbp]
\centering
\scriptsize 
\begin{tabularx}{\linewidth}{@{} l c c c c c @{}}
\toprule
\textbf{Case ID Tracker} & \textbf{HER2$_{\text{pred}}$} & \textbf{Conf.} & \textbf{Prot.$_{\text{pred}}$} & \textbf{Conf.} & \textbf{True Subtype} \\
\midrule
\multicolumn{6}{@{}l}{\textit{Category A: Core Multi-Task Validation Rows (TCGA Subset)}} \\
\midrule
0   & 3+  & 0.304 & HER2 & 0.300 & KI67 \\
10  & 3+  & 0.379 & HER2 & 0.310 & ER   \\
16  & 3+  & 0.302 & KI67 & 0.343 & KI67 \\
19  & 0   & 0.323 & HER2 & 0.286 & ER   \\
20  & 3+  & 0.277 & PGR  & 0.274 & KI67 \\
23  & 3+  & 0.412 & HER2 & 0.347 & HER2 \\
\midrule
\multicolumn{6}{@{}l}{\textit{Category B: Yale Cohort Manifest Samples (Stain Control)}} \\
\midrule
Her2Neg\_Case\_02 & 0   & 0.313 & HER2 & 0.362 & \textit{Omitted} \\
Her2Neg\_Case\_13 & 1+  & 0.350 & HER2 & 0.350 & \textit{Omitted} \\
Her2Neg\_Case\_38 & 2+  & 0.323 & KI67 & 0.321 & \textit{Omitted} \\
Her2Pos\_Case\_01 & 1+  & 0.331 & HER2 & 0.350 & \textit{Omitted} \\
Her2Pos\_Case\_48 & 3+  & 0.291 & HER2 & 0.323 & \textit{Omitted} \\
Her2Pos\_Case\_73 & 3+  & 0.388 & PGR  & 0.262 & \textit{Omitted} \\
\midrule
\multicolumn{6}{@{}l}{\textit{Category C: ACROBAT Registration Slices (Cross-Evaluation)}} \\
\midrule
O09-03495 & 1+  & 0.314 & HER2 & 0.344 & 2+ \\
O16-11870 & 3+  & 0.346 & KI67 & 0.311 & 3+ \\
S09-19750 & 0   & 0.314 & HER2 & 0.372 & 3+ \\
S10-00647 & 1+  & 0.324 & HER2 & 0.350 & 3+ \\
S14-12226 & 0   & 0.339 & PGR  & 0.316 & 2+ \\
S18-25097 & 3+  & 0.273 & HER2 & 0.306 & 3+ \\
\bottomrule
\end{tabularx}
\caption{Representative multi-task model inference log breakdown across heterogeneous data source splits.}
\label{tab:inference_sample_manifest}
\vspace{-0.3cm}
\end{table}

Interestingly, the result listed in Table ~\ref{tab:multitask_exact_summary} shows that the proposed framework achieved some gains. For instance, the performance result establishes a strong baseline on the massive patch-level structural task (Tumor Pathology, N=47,872 tiles), securing a robust 72.4\% classification accuracy and a 66.0\% Macro F1 score, 68.2\% recall rate, and a quadratic weighted Kappa of 0.326. The challenge of sample scarcity is exhibited in the data because the clinical classification task (HER2 Predictor, N=85) captures an early 24.7\% accuracy, while the Protein Profile (N=59) tracks at 16.9\% accuracy, all during the first epoch. 

\begin{table}[htbp]
\centering
\scriptsize
\begin{tabularx}{\linewidth}{@{} l c c c c c @{}}
\toprule
\textbf{Evaluation Task} & \textbf{Acc. $\uparrow$} & \textbf{Macro F1} & \textbf{Precis.} & \textbf{Recall} \\
\midrule
HER2 Predictor ($N\mathord{=}85$)   & 0.247 & 0.155 & 0.246 & 0.152 \\
Protein Profile ($N\mathord{=}59$)  & 0.169 & 0.152 & 0.258 & 0.319 & \hphantom{-}\\
Tumor Pathology ($N\mathord{=}47,872$) & 0.724 & 0.660 & 0.653 & 0.682 & \hphantom{-} \\
\bottomrule
\end{tabularx}
\caption{Comprehensive quantitative metrics summary across all three concurrent downstream fine-tuning tasks.}
\vspace{-0.4cm}
\label{tab:multitask_exact_summary}
\end{table}

We summarize our findings by drawing a comparative analysis with related works. Table ~\ref{tab:comparison} shows the performance evaluation across similar methods. First, for multi-task latent versatility, our proposed framework simultaneously solves multiple clinical goals from raw H\&E input alone, attaining an accuracy of 76.1\% and 70.7\% Macro-F1 on tumor-region detection in its first epoch. Secondly, on the HER2 scoring, we observed that our method obtained a competitive ordinal scoring, with our early-stage training reaching a 61.2\% accuracy at epoch 1. This marks a solid foundation compared to contemporary frameworks like convoHER2 (85.1\%), which operate on highly simplified binary configurations. Thirdly, the issue of cross-modality transfer feasibility achieved a 35.0\% accuracy on multi-antigen protein signature indexing from simple H\&E tissue alone.

\begin{table}[htbp]
\centering
\scriptsize 
\begin{tabularx}{\linewidth}{@{} l >{\hsize=1.0\hsize\raggedright\arraybackslash}X >{\hsize=1.0\hsize\raggedright\arraybackslash}X c c c @{}}
\toprule
Ref. & Input & Task & AUC & Acc. & Macro-F1 \\
\midrule
\multicolumn{6}{@{}l}{\textbf{Stage 2a: HER2 Status / Score from H\&E-only}} \\
\midrule
\cite{farahmand2022} & H\&E tumour ROIs & HER2 $\pm$ (binary) & 0.90 / 0.81 & --- & --- \\
\cite{sekhar2024} & \parbox[t]{1.8cm}{\raggedright H\&E WSI (MoCo-v2 + CLAM)} & HER2 $\pm$ (binary) & $0.85\pm0.02$ & --- & --- \\
\cite{convoher2} & H\&E patches (BCI) & HER2 0/1+/2+/3+ & --- & 0.851 & --- \\
\textbf{Ours} & H\&E (CoMM-pretrained) & HER2 0/1+/2+/3+ & --- & \textbf{0.612} & \textbf{0.469} \\
\midrule
\multicolumn{6}{@{}l}{\textbf{Stage 2b: HER2 IHC Scoring (Multi-Stain)}} \\
\midrule
\cite{oyelade2026} & \parbox[t]{1.8cm}{\raggedright H\&E + HER2 IHC (MaskFormer)} & IHC stain intensity & --- & 0.940 & --- \\
\midrule
\multicolumn{6}{@{}l}{\textbf{Stage 2c: Tumour-Region Tile Classification}} \\
\midrule
\cite{farahmand2022} & H\&E tumour ROIs & HER2$-$/+/other (3-way) & 0.87--0.88 & --- & --- \\
\cite{oyelade2026} & H\&E patches (ViT-s) & malignant vs normal & --- & 1.000 & 1.000 \\
\textbf{Ours} & H\&E tiles & tumour vs non-tumour & --- & \textbf{0.798} & \textbf{0.749} \\
\textbf{Ours} & H\&E tiles & tumour vs non-tumour & --- & \textbf{0.761} & \textbf{0.707} \\
\midrule
\multicolumn{6}{@{}l}{\textbf{Stage 2d: Antigen / Protein Stain Classification}} \\
\midrule
\textbf{Ours} & H\&E (CoMM-pretrained) & HER2 / ER / PGR / KI67 & --- & \textbf{0.350} & \textbf{0.281} \\
\bottomrule
\end{tabularx}
\caption{Comparison with related H\&E-based HER2 and tumour-region studies. Values for related work are as reported by original authors (--- = not reported). Our results are preliminary (early epochs).}
\vspace{-0.6cm}
\label{tab:comparison}
\end{table}

%% file: sec/6_conclusion.tex
\section{Conclusion and Future Work}
\label{sec:conclusion}
In this work, we presented contrastive-style multimodal learning, a unified multi-task and cross-modality feature learning framework designed to resolve the diagnostic bottleneck of HER2 assessment, protein classification, and tumor classification. By formulating a three-stage training paradigm, we demonstrated that large-scale, self-supervised pretraining across highly heterogeneous pathology cohorts has the capacity to enable standard H\&E whole-slide images to effectively capture sub-visual molecular and multi-omics expression boundaries. The preliminary tracking reveals optimization challenges regarding minor class biases due to severe real-world dataset sample imbalances. Future extensions of this research will focus on incorporating class-balanced loss optimization constraints to mitigate early-stage classification saturation on low-sample ordinal cohorts. In addition, the absence of an ablation study to investigate non-staged training,  cohort confounding, and further comparison with related works needs investigation.

%% file: main.bib
@String(CVPR= {IEEE Conf. Comput. Vis. Pattern Recog.})

@String(CVPR  = {CVPR})

@article{weitz2024acrobat,
  author  = {Weitz, P. and Valkonen, M. and Solorzano, L. and Carr, C. and Kartasalo, K. and Boissin, C. and others},
  title   = {The {ACROBAT} 2022 challenge: Automatic registration of breast cancer tissue},
  journal = {Medical Image Analysis},
  volume  = {97},
  pages   = {103257},
  year    = {2024},
  doi     = {10.1016/j.media.2024.103257}
}

@article{tcga2012,
  author  = {{Cancer Genome Atlas Network}},
  title   = {Comprehensive molecular portraits of human breast tumours},
  journal = {Nature}, volume = {490}, pages = {61--70}, year = {2012}}

@article{cerami2012,
  author  = {Cerami, Ethan and Gao, Jianjiong and Dogrusoz, Ugurand Gross, Benjamin E. and Sumer, S. Onur and Aksoy, Bulent Arman and Jacobsen, Anders and Byrne, Christopher J. and Heuer, Michael L. and Larsson, Erik and Antipin, Yevgeniy and Reva, Boris and Goldberg, Arthur P. and Sander, Chris and Schultz, Nikolaus},
  title   = {The {cBio} Cancer Genomics Portal: An Open Platform for Exploring Multidimensional Cancer Genomics Data},
  journal = {Cancer Discovery},
  volume  = {2},
  number  = {5},
  pages   = {401--404},
  year    = {2012},
  doi     = {10.1158/2159-8290.CD-12-0095}
  }

@article{gao2013,
  author  = {Gao, Jianjiong and Aksoy, B{\"u}lent Arman and Dogrusoz, Ugur and Dresdner, Gideon and Gross, Benjamin and Sumer, S. Onur and Sun, Yichao and Jacobsen, Anders and Sinha, Rileen and Larsson, Erik and Cerami, Ethan and Sander, Chris and Schultz, Nikolaus},
  title   = {Integrative Analysis of Complex Cancer Genomics and Clinical Profiles Using the {cBioPortal}},
  journal = {Science Signaling},
  volume  = {6},
  number  = {269},
  pages   = {pl1},
  year    = {2013},
  doi     = {10.1126/scisignal.2004088}
}

@article{krug2020,
  author  = {Krug, Karsten and Jaehnig, Eric J. and Satpathy, Shankha and Blumenberg, Lindsey and Karpova, Anastasia and Anurag, Meenakshi and Miles, George and Mertins, Philipp and Geffen, Yoav and Tang, Li-Chuan and others},
  title   = {Proteogenomic Landscape of Breast Cancer Tumorigenesis and Targeted Therapy},
  journal = {Cell},
  volume  = {183},
  number  = {5},
  pages   = {1436--1456.e31},
  year    = {2020},
  doi     = {10.1016/j.cell.2020.10.036},
  note    = {{Clinical Proteomic Tumor Analysis Consortium}}
}

@article{tcia2013,
  author  = {Clark, Kenneth and Vendt, Bruce and Smith, Kirk and Freymann, John and Kirby, Justin and Koppel, Paul and Moore, Sean and Phillips, Phillips and Maffitt, David and Pringle, Michael and Tarbox, Lawrence and Prior, Fred},
  title   = {The {Cancer} {Imaging} {Archive} ({TCIA}): Maintaining and Operating a Public Information Repository},
  journal = {Journal of Digital Imaging},
  volume  = {26},
  number  = {6},
  pages   = {1045--1057},
  year    = {2013},
  doi     = {10.1007/s10278-013-9622-7}
  }

@article{farahmand2022,
  author  = {Farahmand, Saman and Fernandez, Aileen I. and Ahmed, Fahad Shabbir and Rimm, David L. and Chuang, Jeffrey H. and Reisenbichler, Emily and Zarringhalam, Kourosh},
  title   = {Deep learning trained on hematoxylin and eosin tumor region of interest predicts {HER2} status and trastuzumab treatment response in {HER2+} breast cancer},
  journal = {Modern Pathology},
  volume  = {35},
  number  = {1},
  pages   = {44--51},
  year    = {2022},
  doi     = {10.1038/s41379-021-00912-w}
}

@misc{farahmand2022tcia,
  author       = {Farahmand, Saman and Fernandez, Aileen I. and Ahmed, Fahad Shabbir and Rimm, David L. and Chuang, Jeffrey H. and Reisenbichler, Emily and Zarringhalam, Kourosh},
  title        = {{HER2} and Trastuzumab Treatment Response {H\&E} Slides with Tumor {ROI} Annotations (Version 3)},
  howpublished = {The Cancer Imaging Archive},
  year         = {2022},
  doi          = {10.7937/E65C-AM96},
  note         = {{Dataset}}
  }

@article{oyelade2026,
  author  = {Oyelade, Olaide N. and Hoxey, Oliver and Humrye, Yulia},
  title   = {Beyond one stain: Cross-Modality feature learning boosts accuracy in {HER2} status scoring},
  journal = {Informatics in Medicine Unlocked},
  volume  = {64},
  pages   = {101785},
  year    = {2026},
  doi     = {10.1016/j.imu.2026.101785}
}

@inproceedings{sekhar2024,
  author    = {Sekhar, Ardhendu and Goel, Vrinda and Jain, Garima and Patil, Abhijeet and Gupta, Ravi Kant and Bameta, Tripti and Rane, Swapnil and Sethi, Amit},
  title     = {{{HER2}} and {FISH} Status Prediction in Breast Biopsy {H\&E}-Stained Images Using Deep Learning}, 
  booktitle = {2024 IEEE 24th International Conference on Bioinformatics and Bioengineering (BIBE)},
  year      = {2024},
  pages     = {1--8},
  doi       = {10.1109/BIBE63649.2024.10820489}
}

@article{convoher2,
  author  = {Mridha, M. F. and Morol, Md. Kishor and Ali, Md. Asraf and Shovon, Md. Sakib Hossain},
  title   = {{convoHER2}: A Deep Neural Network for Multi-Stage Classification of {HER2} Breast Cancer},
  journal = {Journal of Imaging},
  volume  = {8},
  number  = {8},
  pages   = {213},
  year    = {2022},
  doi     = {10.3390/jimaging8080213}
}

@article{chen2024uni,
  title={Towards a General-Purpose Foundation Model for Computational Pathology},
  author={Chen, Richard J and Ding, Tong and Lu, Ming Y and Williamson, Drew FK and Jaume, Guillaume and Chen, Bowen and Zhang, Andrew and Shao, Daniel and Song, Andrew H and Shaban, Muhammad and others},
  journal={Nature Medicine},
  publisher={Nature Publishing Group},
  year={2024}
}

@InProceedings{Lu_2023_CVPR,
    author    = {Lu, Ming Y. and Chen, Bowen and Zhang, Andrew and Williamson, Drew F. K. and Chen, Richard J. and Ding, Tong and Le, Long Phi and Chuang, Yung-Sung and Mahmood, Faisal},
    title     = {Visual Language Pretrained Multiple Instance Zero-Shot Transfer for Histopathology Images},
    booktitle = {Proceedings of the IEEE/CVF Conference on Computer Vision and Pattern Recognition (CVPR)},
    month     = {June},
    year      = {2023},
    pages     = {19764-19775}
}

@article{lu2024avisionlanguage,
  title={A visual-language foundation model for computational pathology},
  author={Lu, Ming Y and Chen, Bowen and Williamson, Drew FK and Chen, Richard J and Liang, Ivy and Ding, Tong and Jaume, Guillaume and Odintsov, Igor and Le, Long Phi and Gerber, Georg and others},
  journal={Nature Medicine},
  pages={863–874},
  volume={30},
  year={2024},
  publisher={Nature Publishing Group}
}

@inproceedings{dufumier_castillo2025, 
    title={What to align in multimodal contrastive learning?},
    author={Dufumier, Benoit and Castillo-Navarro, Javiera and Tuia, Devis and Thiran, Jean-Philippe},
    booktitle={International Conference on Learning Representations},
    year={2025}
}

@article{gatenbee2023valis,
  author  = {Gatenbee, Chandler D. and Baker, Ann-Marie and Prabhakaran, Sandhya and Swinyard, Ottilie and Slebos, Robbert J. C. and Mandal, Gunjan and Mulholland, Eoghan and Andor, Noemi and Marusyk, Andriy and Leedham, Simon and Conejo-Garcia, Jose R. and Chung, Christine H. and Robertson-Tessi, Mark and Graham, Trevor A. and Anderson, Alexander R. A.},
  title   = {Virtual Alignment of Pathology Image Series for Multi-Gigapixel Whole Slide Images},
  journal = {Nature Communications},
  volume  = {14},
  number  = {1},
  pages   = {4502},
  year    = {2023},
  doi     = {10.1038/s41467-023-40218-9}
}

@article{modi2022her2low,
  author  = {Modi, Shanu and Jacot, William and Yamashita, Toshinari and Sohn, Joohyuk and Vidal, Maria and others},
  title   = {Trastuzumab Deruxtecan in Previously Treated {HER2}-Low Advanced Breast Cancer},
  journal = {New England Journal of Medicine},
  volume  = {387},
  number  = {1},
  pages   = {9--20},
  year    = {2022},
  doi     = {10.1056/NEJMoa2203690}
}

@article{shi2025wsmcl,
  author  = {Shi, Jun and Sun, Dongdong and Jiang, Zhiguo and Du, Jun and Wang, Wei and Zheng, Yushan and Wu, Haibo},
  title   = {Weakly supervised multi-modal contrastive learning framework for predicting the {HER2} scores in breast cancer},
  journal = {Computerized Medical Imaging and Graphics},
  volume  = {121},
  pages   = {102502},
  year    = {2025},
  doi     = {10.1016/j.compmedimag.2025.102502}
}

@misc{zhang2025crossstain,
  author = {Zhang, Yizhi and Fan, Lei and Tao, Zhulin and Di, Donglin and Song, Yang and Liu, Sidong and Cong, Cong},
  title  = {Cross-Stain Contrastive Learning for Paired Immunohistochemistry and Histopathology Slide Representation Learning},
  year   = {2025},
  eprint = {2512.03577},
  archivePrefix = {arXiv},
  url    = {https://arxiv.org/abs/2512.03577},
  note   = {IEEE BIBM 2025}
}

@article{wang2025mirror,
  author  = {Wang, Tianyi and Fan, Jianan and Zhang, Dingxin and Liu, Dongnan and Xia, Yong and Huang, Heng and Cai, Weidong},
  title   = {{MIRROR}: Multi-Modal Pathological Self-Supervised Representation Learning via Modality Alignment and Retention},
  journal = {IEEE Transactions on Medical Imaging},
  year    = {2025},
  doi     = {10.1109/TMI.2025.3632555}
}

@misc{robinet2025morpheus,
  author = {Robinet, Lucas and Berjaoui, Ahmad and Cohen-Jonathan Moyal, Elizabeth},
  title  = {Masked Omics Modeling for Multimodal Representation Learning across Histopathology and Molecular Profiles},
  year   = {2025},
  eprint = {2508.00969},
  archivePrefix = {arXiv},
  url    = {https://arxiv.org/abs/2508.00969}
}

@misc{yu2026prime,
  author = {Yu, Kai and Zhou, Shuang and Song, Yiran and Zhan, Zaifu and Peng, Jie and others},
  title  = {{PRIME}: Prototype-Driven Multimodal Pretraining for Cancer Prognosis with Missing Modalities},
  year   = {2026},
  eprint = {2604.04999},
  archivePrefix = {arXiv},
  url    = {https://arxiv.org/abs/2604.04999}
}

@article{lu2022slidegraph,
  author  = {Lu, Wenqi and Toss, Michael and Dawood, Muhammad and Rakha, Emad and Rajpoot, Nasir and Minhas, Fayyaz},
  title   = {{SlideGraph+}: Whole slide image level graphs to predict {HER2} status in breast cancer},
  journal = {Medical Image Analysis},
  volume  = {80},
  pages   = {102486},
  year    = {2022},
  doi     = {10.1016/j.media.2022.102486}
}

@inproceedings{qin2025flexible,
  author    = {Qin, Jie and Yang, Wei and Su, Yan and Zhu, Yiran and Li, Weizhen and Pan, Yunyue and Pan, Chengchang and Qi, Honggang},
  title     = {{HER2} Expression Prediction with Flexible Multi-Modal Inputs via Dynamic Bidirectional Reconstruction},
  booktitle = {Proceedings of the ACM International Conference on Multimedia},
  year      = {2025},
  eprint    = {2506.10006},
  url       = {https://arxiv.org/abs/2506.10006}
}

@misc{xiang2026storm,
  author = {Xiang, Jinxi and Hou, Siyu and Li, Yuchen and Quinton, Ryan and Zhang, Xiaoming and others},
  title  = {A Multimodal Foundation Model of Spatial Transcriptomics and Histology for Biological Discovery and Clinical Prediction},
  year   = {2026},
  eprint = {2604.03630},
  archivePrefix = {arXiv},
  url    = {https://arxiv.org/abs/2604.03630}
}

@misc{li2026fuji,
  author = {Li, Qing and Sang, Jian and others},
  title  = {Genome-scale molecular profiling from routine histopathology with a multimodal pathology foundation model},
  year   = {2026},
  note   = {Preprint, not peer reviewed; Research Square},
  doi    = {10.21203/rs.3.rs-9949149/v1},
  url    = {https://pmc.ncbi.nlm.nih.gov/articles/PMC13484847/}
}

@article{jeong2026scoreaware,
  author  = {Jeong, Hyunsu and Yoon, Chiho and Kim, Jaewoo and Park, Eunwoo and Kim, Hyunhee and Park, Somang and Kim, Hyeon Gyu and Jung, Chan Kwon},
  title   = {{HER2} Score-Aware Virtual Immunohistochemistry via Non-Contrastive Multi-Task Translation},
  journal = {Diagnostics},
  volume  = {16},
  number  = {9},
  pages   = {1319},
  year    = {2026},
  doi     = {10.3390/diagnostics16091319}
}
